\pdfoutput=1

\documentclass[11pt]{article}
\usepackage{emnlp2021}

\usepackage{times}
\usepackage{latexsym}

\usepackage[T1]{fontenc}

\usepackage[utf8]{inputenc}

\usepackage{microtype}

\usepackage{amsmath,amssymb,amsfonts}
\usepackage{booktabs}
\usepackage{graphicx}
\usepackage{subcaption}
\usepackage{algorithm} 
\usepackage{algpseudocode}
\usepackage{multirow}
\usepackage{makecell}
\usepackage{hyperref}
\usepackage{color}
\newcommand*{\affaddr}[1]{#1} 
\newcommand*{\affmark}[1][*]{\textsuperscript{#1}}
\newcommand*{\email}[1]{\texttt{#1}}

\title{Accurate, yet inconsistent? Consistency Analysis on Language Understanding Models}

\author{
Myeongjun Jang\affmark[1]\thanks{equal contribution}, 
Deuk Sin Kwon\affmark[2]\footnotemark[1] \and
Thomas Lukasiewicz\affmark[1]\\
\affaddr{\affmark[1]Department of Computer Science, University of Oxford}\\
\email{\texttt{\{myeongjun.jang, thomas.lukasiewicz\}@cs.ox.ac.uk}}\\
\affaddr{\affmark[2]Language Super Intelligence Labs, SK Telecom}\\
\email{kds0281@gmail.com}
}

\date{}

\begin{document}
\maketitle

\begin{abstract}
Consistency, which refers to the capability of generating the same predictions for semantically similar contexts, is a highly desirable property for a sound language understanding model. Although recent pretrained language models (PLMs) deliver outstanding performance in various downstream tasks, they should exhibit consistent behaviour provided the models truly understand language. In this paper, we propose a simple framework named \emph{\textbf{c}onsistency \textbf{a}nalysis on \textbf{l}anguage \textbf{u}nderstanding \textbf{m}odels (CALUM)} to evaluate model's lower-bound consistency ability. Through experiments, we confirmed that current PLMs are prone to generate inconsistent predictions even for semantically identical inputs. We also observed that multi-task training with paraphrase identification tasks is of benefit to improve consistency, increasing the consistency by 13\% on average.
\end{abstract}

\section{Introduction}
Large-sized pretrained language models (PLMs), such as BERT \cite{BERT} and GPT2 \cite{GPT2}, have entered the mainstream in natural language processing (NLP). They delivered outstanding performance on many downstream tasks through fine-tuning and in-context learning \cite{GPT3}. Based on these results, claims that such PLMs understand language have been introduced in the literature \cite{BERT, ohsugi2019simple, qiu2020pre} and popular press, such as \textit{Google Blog} post\footnote{\href{https://www.blog.google/products/search/search-language-understanding-bert/}{https://www.blog.google/products/search/search-language-understanding-bert/} accessed 2021/07/14} and \textit{Toward Data Science} website\footnote{\href{https://towardsdatascience.com/pre-trained-language-models-simplified-b8ec80c62217}{https://towardsdatascience.com/pre-trained-language-models-simplified-b8ec80c62217} accessed 2021/07/14}.

However, recent studies cast doubts on the language understanding capacity of PLMs. Nuemerous studies revealed that PLMs are incapable of capturing the semantic meaning of sentences but rely on the excessive exploitation of statistical cues or syntactic patterns \cite{habernal2018argument, niven2019probing, mccoy2019right, benderkoller2020climbing}. Another line of works found that PLMs memorise frequent word/phrase/knowledge presented in pretraining data but hardly understand unseen expressions and knowledge \cite{kassner2020pretrained, ravichander2020systematicity, hofmann2021superbizarre}. 
Moreover, many works observed that PLMs lack understanding of negated phrase \cite{naik2018stress, hossain2020analysis, kassner2020negated, ettinger2020bert, hosseini2021understanding} or number-related commonsense knowledge \cite{lin2020birds}.

The lexical definition of the word \textit{understand} is \textit{to know or realise the meaning of words and a language}\footnote{\href{https://www.oxfordlearnersdictionaries.com/definition/english/understand?q=understand}{https://www.oxfordlearnersdictionaries.com/definition-/english/understand?q=understand} accessed 2021/07/14}. In this regard, \textit{consistency}, which refers to the capability of making coherent decisions on semantically equivalent contexts \cite{elazar2021measuring}, is an important property that a good language understanding model should satisfy. Language models should exhibit consistent behaviour provided they truly understand human language. In this context, the performance of PLMs should be illuminated and evaluated in terms of consistency, aside from other evaluation metrics such as accuracy.

Many recent studies deliver indications that PLMs lack consistency. In a zero-shot knowledge retrieval task, BERT made different predictions for semantically similar queries ; queries that subjects are replaced by their plural forms \cite{ravichander2020systematicity} or paraphrased queries \cite{elazar2021measuring}. Research regarding text adversarial attacks also revealed that fine-tuned PLMs misclassify adversarial examples where human-indiscernible noises are added to the original text \cite{morris2020textattack, li2020bertattack, gargramakrishnan2020bae}.

In this paper, we propose a framework named \textbf{C}onsistency \textbf{A}nalysis on \textbf{L}anguage \textbf{U}nderstanding \textbf{M}odels (CALUM) to evaluate the consistency of PLMs after fine-tuned on language understanding tasks. The framework measures consistency by leveraging marginal perturbations that guarantee perfectly identical semantic meanings. The major contributions of our work could be summarised as follows:

\begin{itemize}
    \item We propose a framework for evaluating PLM's lower-bound ability of consistency on language understanding tasks.
    \item Through experiments, we reveal that PLMs are inconsistent even for input text conveying identical semantic meanings regardless of models and languages.
    \item Our finding suggests that training a multi-task model with semantic textual similarity tasks can improve consistency.
\end{itemize}

\section{Related Works}
\paragraph{Consistency} There have been several attempts to analyse the consistency of language models in various NLP domains. For zero-shot knowledge retrieval tasks, \citet{ravichander2020systematicity} found that PLMs generate different answers if an object of the original query is replaced with its plural form (e.g. `A robin is a [MASK]' to `robins are [MASK]'). \citet{elazar2021measuring} observed discrepancy in the predictions of PLMs for paraphrase queries and alleviated the issue by fine-tuning the model on the generated paraphrase queries. In question answering (QA), \citet{ribeiro-etal-2019-red} showed that state-of-the-art QA and visual QA models generate inconsistent outputs for queries with the same context. They used data augmentation to improve consistency. To train a consistent QA model, \citet{asaihajishirzi2020logic} also used data augmentation and, additionally, inconsistency loss, which is designed to penalise inconsistent predictions. \citet{ribeiro2020beyond} proposed Invariance test to evaluate consistency. For a sentiment analysis task, they changed the named entity presented in a given sentence because such perturbation does not change the polarity of the sentence. However, it is difficult to apply this approach to other language understanding tasks because it requires human inspection since the method does not guarantee semantic equivalence. For instance, "How many people are there in London?" and "How many people are there in Seoul?" convey entirely different meaning and, therefore, there exist the possibility for a model to make different decisions. Research on consistency in other domains includes text summarisation \cite{kryscinski2020evaluating}, explanation generation \cite{camburu-etal-2020-make}, and dialog generation \cite{li2020dont}. \citet{li2020dont} employed unlikelihood training \cite{welleck2019neural} to improve consistency of a dialog model.

\paragraph{Text Adversarial Attack} After the advent of PLMs, there have been several works to attack the models by using adversarial examples. \citet{jin2020bert} proposed a black-box attack approach called \textsc{TextFooler} that replace important words in an input sentence with their synonyms. \citet{li2020bertattack} used BERT for generating adversarial samples. They first extracted important words for decision-making and replaced them by using BERT masked language model (MLM). \citet{gargramakrishnan2020bae} also used BERT MLM not only for replacing important tokens but also for inserting new tokens. \citet{li2021contextualized} employed MLM for three strategies; replace, insert, and merge that mask bi-gram and replace it with a single word. 

Text adversarial attacks have a commonality with consistency in that adversarial examples are designed to have a similar meaning with their original counterparts. However, text adversarial attacks exhibit a critical difference to our work in that they do not guarantee semantic identicalness due to the unnatural looking, grammatical errors, and insecurity of PLM's MLM performance \cite{ravichander2020systematicity, ettinger2020bert, kassner2020negated, elazar2021measuring}. On the contrary, our framework perfectly assures identical meaning, which is an essential property to evaluate a lower-bound consistency of PLMs.

\paragraph{Prompt Engineering} \textit{Prompt} refers to constructing an input text with a job description or some examples prepended \cite{lester2021power}. Recently, there has been an upsurge of research interest in prompt engineering, with the advent of GPT3 showing its powerful performance on zero-shot/few-shot tasks only through variants of prompts without further fine-tuning \cite{brown2020language}. Considering how to effectively search prompts that optimally control the fixed pre-trained language model, several studies have been conducted on automating prompt design or tuning the prompt with learnable parameters \cite{shin2020autoprompt, reynolds2021prompt}. \citet{lester2021power} proposed a prompt tuning method where some tunable tokens presenting each downstream task are prepeneded to the original input text with the frozen pre-trained model, showing its competitiveness with a typical fine-tuning method. \citet{li2021prefix} also suggested a novel approach for prompt engineering  called \textit{prefix-tuning}, which prepends the task-specific prefix consisting of tunable parameters to the inputs in each transformer layer and optimises only the prefix instead of updating all Transformer parameters. In line with these prompt engineering techinques, \citet{hambardzumyan2021warp} purturbed inputs as a way of Adversarial Reprogramming by inserting learnable prompt tokens to a specific location in the input text, training the prompt parameters during optimization of the specific downstream task objective.

However, these studies which highlight different approaches of prompt engineering and subsequent variations of performance, raise questions about whether language models have consistency capability. In addition, although the methods proposed in these works find the optimal for solving downstream tasks through prompt tuning, there are still doubts on whether the consistency of LMs is improved by these methods. Moreover, a majority of these studies focus only on the improvement of the performance of downstream tasks. Therefore, there certainly is a need for evaluating the language understanding ability of LMs from a consistency point of view. 

\section{CALUM: Consistency Analysis on Language Understanding Models}
\begin{figure}[t!]
	\centering
	\begin{subfigure}[b]{0.5\textwidth}
		\includegraphics[width=\linewidth]{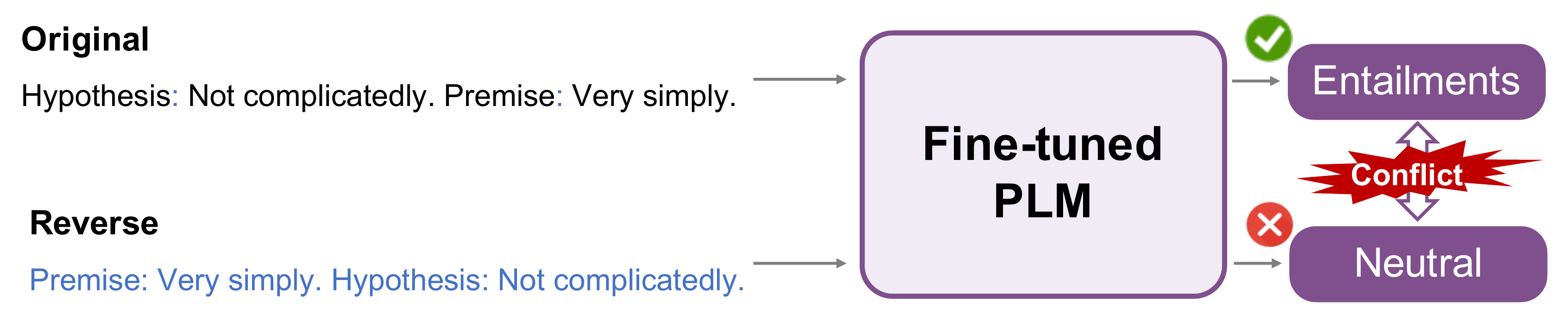}
		\caption{\textsc{REVERSE} case} \label{fig:reverse}
	\end{subfigure} \
	\begin{subfigure}[b]{0.5\textwidth}
		\includegraphics[width=\linewidth]{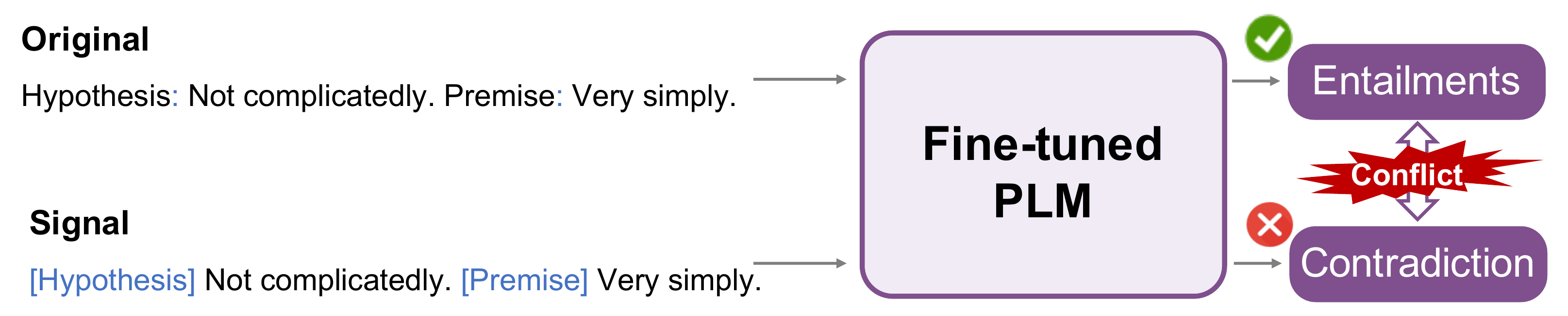}
		\caption{\textsc{SIGNAL} case} \label{fig:signal}
	\end{subfigure}%
	\caption{Example of CALUM framework for MNLI task. The changes in the original free-text inputs are marked in blue.}
	\label{fig:framework}
\end{figure}
Our framework evaluates a model's consistency on language understanding tasks that infer the relation of two input sentences, such as natural language inference (NLI) and semantic textual similarity (STS) tasks. In our experiments, it is crucial to ensure semantic equivalence after perturbation. To achieve this, we first insert a free-text sentence-type indicator at the beginning of each sentence (e.g. `Premise:' and `Hypothesis:' for the NLI task). Each model is trained on this original form. Next, we applied two perturbation methods: \textsc{REVERSE} and \textsc{SIGNAL}.

\paragraph{\textsc{REVERSE}} This method changes the order of the two input sentences. The meaning of inputs remains identical because of the sentence-type indicator. An example case of this method is illustrated in Figure \ref{fig:reverse}.

\paragraph{\textsc{SIGNAL}} This method changes a special symbol in the sentence-type indicator. In our experiments, we replaced a colon with brackets. This ensures semantic equivalence because such symbols convey no specific meanings. An example case of this mehod is illustrated in Figure \ref{fig:signal}.

Finally, we measure a model's consistency by calculating the accuracy between its predictions on the original and perturbed inputs. The overall framework of our proposed method is illusrtated in Figure \ref{fig:framework}.

\section{Experiment Design}

\subsection{Datasets}
 As the two task types (i.e. NLI \& STS) are mentioned in the previous section, we select the related tasks from the GLUE Benchmark \cite{wang2018glue}. For the NLI task, we use \textbf{MNLI} (MultiNLI, Multi-Genre Natural Language Inference, \citealp{williams-etal-2018-broad}), \textbf{QNLI} (Question Natural Language Inference, \citealp{rajpurkar-etal-2016-squad}) and \textbf{RTE} (Recognising Textual Entailment, \citealp{candela2006evaluating}) for the evaluation of our framework; they are composed of two sentence pairs and a label indicating whether the sentence pairs are entailed or not in common. In addition, we also use \textbf{QQP} (Quora Question Pairs\footnote{\href{https://www.kaggle.com/c/quora-question-pairs/data}{https://www.kaggle.com/c/quora-question-pairs/data}}) and \textbf{MRPC} (Microsoft Research Paraphrase Corpus,  \citealp{dolan2005automatically}) for the STS task type, which  consist of two sentence pairs and a label related to sentence equivalence.

Meanwhile, in order to show the general applicability of our framework, we perform the evaluation not only on the above English tasks, but also on Korean versions of the same task type. For the NLI task, \textbf{KorNLI} \cite{ham2020kornli} \& \textbf{KLUE-NLI} \cite{park2021klue} are selected and for the STS task, \textbf{KLUE-STS} \cite{park2021klue} is used. The three Korean datasets do not provide test sets. Therefore, we randomly sampled test sets from the validation set for \textbf{KLUE} datsets and from training set for \textbf{KorNLI} dataset. Descriptions of the datasets are illustrated in Table \ref{table.dataset}.

\begin{table}[t!]
	\begin{center}
		\caption{Descriptions of datasets for the experiments.} \label{table.dataset}%
		\renewcommand{\arraystretch}{1.7}
		\footnotesize{
			\centering{\setlength\tabcolsep{3pt}
				\begin{tabular}{ccccc}
					\toprule
					\hline
					& \makecell[c]{\# of \\ classes} & \makecell[c]{Train set \\ size} &
					\makecell[c]{Validation \\ set size} &
					\makecell[c]{Test set \\ size} \\
					\hline
					MNLI & 3 & 393K & 9.8K & 9.8K \\
					QNLI & 2 & 105K & 5.5K & 5.5K \\
					RTE & 2 & 2.5K & 277 & 3K \\
					QQP & 2 & 364K & 40K & 391K \\
					MRPC & 2 & 3.7K & 408 & 1.7K \\
					KorNLI & 3 & 53K & 10K & 10K \\
					KLUE-NLI & 3 & 25K & 1.5K & 1.5K \\
					KLUE-STS & 2 & 1.2K & 260 & 259 \\
					\hline
					\bottomrule	
		\end{tabular}}}
	\end{center}
\end{table}

\subsection{Model candidates}
To measure the consistency of PLMs on language understanding capability, we conduct experiments on various types and sizes of PLMs. 
For English tasks, we perform experiments with the Encoder models ($RoBERTa_{base}$, $RoBERTa_{large}$, $Electra_{small}$, $Electra_{large}$), the Decoder models ($GPT2_{base}$, $GPT2_{large}$), and the Seq2Seq models ($T5_{base}$, $T5_{large}$, $BART_{base}$). 
For Korean tasks, $KoBERT$ and $KoElectra$ are used as the Encoder models. For the Decoder and Seq2Seq models, $KoGPT2$, and $KoBART$ are employed, respectively. All PLMs implemented in our experiments are from huggingface transformers \cite{wolf2020transformers} library and used as fine-tuning backbone models.
\subsection{Training details}\label{section.training_detail}
Regardless of the model type except for $T5$ models , a classification head is added on the top of the model, and all weights of the model are updated while optimising the classification objective function. We finetune each of our candidate language models on individual task datasets.  Meanwhile, finetuning for $T5$ models is not performed in this study since the finetuned version of T5 models is used.

At finetuning, we use the AdamW optimiser with weight decay 1\textit{e}-3, \cite{loshchilov2018fixing} and a linear learning rate scheduler with warm-up. We finetune models for 10 epochs with learning rate of 1\textit{e}-5 and batch size of 64, applying an early stopping method during the training.


\subsection{Human Performance}
We also evaluate human's consistency ability. For English and Korean tasks, five human annotators who are native to each language are asked to solve the language understanding tasks we used for the experiments. For each annotator, we provide the original input format. Each annotator is given 30 samples extracted from the validation data, and their corresponding perturbed examples.

\begin{table*}[t!]
	\begin{center}
		\renewcommand{\arraystretch}{1.3}
		\footnotesize{
			\centering{\setlength\tabcolsep{3pt}
		\begin{tabular}{c|ccc|ccc|ccc|ccc|ccc}
		\toprule
		\hline
		\multirow{2}{*}{Model} & 
		\multicolumn{3}{c|}{MNLI} & \multicolumn{3}{c|}{QNLI} &  \multicolumn{3}{c|}{RTE} & 
		\multicolumn{3}{c|}{QQP} &  \multicolumn{3}{c}{MRPC} \\ 
		& $Acc_{val}$ & $C_{R}$ & $C_{S}$ 
		& $Acc_{val}$ & $C_{R}$ & $C_{S}$ 
		& $Acc_{val}$ & $C_{R}$ & $C_{S}$ 
		& $Acc_{val}$ & $C_{R}$ & $C_{S}$ 
		& $Acc_{val}$ & $C_{R}$ & $C_{S}$ 
		\\ \hline
		$RoBERTa_{base}$ 
		& 87.2 & 60.3 & 98.6 
		& 92.4 & 75.5 & 98.5 
		& 66.7 & 66.4 & 92.2 
		& 90.8 & 97.3 & 99.1 
		& 87.0 & 94.9 & 97.7 
		\\
		$RoBERTa_{large}$ 
		& 90.0 & 64.3 & \textbf{98.9} 
		& 94.1 & 81.8 &	\textbf{99.0} 
		& 74.7 & 59.6 & 90.0 
		& 91.2 & 96.7 & \textbf{99.3} 
		& 84.4 & 93.9 & 96.4 
		\\
		$Electra_{small}$ 
		& 81.3 & 52.9 & 94.5 
		& 87.2 & 64.6 & 94.8 
		& 55.4 & 86.6 & 92.0 
		& 88.5 & 97.1 & 97.5 
		& 68.4 & \textbf{100.0} & \textbf{100.0} 
		\\
		$Electra_{large}$ 
		& \textbf{90.7} & 63.3 & 98.6 
		& \textbf{94.7} & 55.4 &	\textbf{99.0} 
		& \textbf{82.2} & 59.2 & 94.8 
		& \textbf{91.4} & 97.1 & 99.1 
		& \textbf{87.7} & 92.6 & 96.4 
		\\
		\hline
		$GPT2_{base}$ 
		& 80.4 & 44.0 & 92.2 
		& 87.4 & 49.0 & 94.1 
		& 61.0 & 62.5 & 83.4 
		& 87.9 & 93.1 & 95.7 
		& 70.1 & 92.2 & 86.8 
		\\
		$GPT2_{large}$ 
		& 85.9 & 56.8 & 96.9 
		& 91.4 & 52.7 &	97.6 
		& 70.4 & 40.1 & 93.7 
		& 90.3 & 94.1 & 98.8 
		& 80.9 & 89.7 & 90.0 
		\\
		$BART_{base}$ 
		& 85.5 & 54.2 & 98.5 
		& 91.4 & 50.2 &	98.5 
		& 58.1 & 68.7 & 91.5 
		& 89.9 & 96.8 & 98.9 
		& 70.8 & 99.8 & 99.2 
		\\
		\hline
		$T5_{base}$ 
		& 85.9 & 60.3 & 6.4 
		& 93.2 & 87.4 & 0.1 
		& 66.4 & 82.1 & 0.0 
		& 90.6 & 97.4 & 63.6 
		& 85.8 & 96.4 & 0.0 
		\\
		$T5_{large}$ 
		& 89.8 & 85.7 & 83.2 
		& 93.9 & 94.5 &	53.7 
		& 79.4 & 87.1 & 0.0 
		& 91.3 & 97.7 & 70.5 
		& \textbf{87.7} & 97.5 & 0.0 
		\\
		\hline
		Human
		& 77.3 & \textbf{96.0} & 96.0 
		& 85.5 & \textbf{98.3} &	98.3 
		& 79.3 & \textbf{94.0} & \textbf{96.0} 
		& 80.5 & \textbf{98.3} & 98.3 
		& 59.3 & 91.9 & \textbf{100.0} 
		\\
		\hline
		\bottomrule
		\end{tabular}}}
	\caption{Results for the consistency evaluation on English datasets. $Acc_{val}$ denotes an accuracy on the validation dataset. $C_{R}$ and $C_{S}$ stands for the consistency for \textsc{REVERSE} and \textsc{SIGNAL} cases, respectively.  We trained each model five times and recorded the average of each metric The best values are in bold.} \label{table.eng_experiment}%
	\end{center}
\end{table*}

\section{Experimental Results}
\subsection{Experiments on English Datasets}
The experimental results for English datasets are summarised in Table \ref{table.eng_experiment}. We repeated the experiment for each model and dataset five times, except for the $T5$ model, since the models provided by HuggingFace are already trained on the five tasks through multi-task training.

In general, the models are highly consistent in \textsc{SIGNAL} case but fall short of expectation in \textsc{REVERSE} case. The only exceptions were $T5_{base}$ and $T5_{large}$ model. On the contrary, humans achieved a high consistency level on both \textsc{REVERSE} and \textsc{SIGNAL}, even though they showed a lower level of accuracy than the models in several tasks. We observed that, although a model generates a high consistency, it is not trustworthy provided the accuracy is significantly low. For instance, $Electra_{small}$ model trained on the MRPC dataset predicts all instances as `equivalent'. Despite 100\% consistency, the model is incapable of understanding language. We also describe several examples of inconsistent predictions in Tables \ref{table.Example_roberta} and \ref{table.example_t5_signal}. More examples are available in Tables \ref{table.Example_electra_large}, \ref{table.Example_t5_large} and \ref{table.example_t5base_signal} in Appendix. Further detailed analyses of the results apart from the $T5$ models are described below. The analysis on the results of $T5$ model is provided separately in section \ref{section:t5_analysis} since they exhibited entirely different results compared to the others. 

\subsubsection{Analysis on Task Types}
For each model, we performed a t-test on the results according to the task type, i.e. NLI and STS. For the \textsc{REVERSE} case, the consistency of STS tasks outperformed that of NLI tasks under a significance level of 0.01. The margin was considerable, and the consistency of STS tasks is almost close to the human performance. A possible reason for these results is that the objective of STS tasks is identifying whether two sentences with different wordings convey the same meaning. Therefore, models trained on such tasks can capture the intrinsic meaning of sentences better and thus, more robust to the meaning-preserving perturbations. Our results are colinear with the work of \cite{elazar2021measuring} that improved a model's consistency by paraphrase data augmentation.

For the \textsc{SIGNAL} case, however, the difference of consistency between NLI and STS tasks was marginal. Instead, the number of training data played a more important role in improving consistency. The consistency performance of the RTE and MRPC tasks is worse than the others, and the difference was statistically significant with a significance level of 0.01. From these results, we gained an insight that STS tasks and the number of training data are important contributing factors to the improvement of consistency.

\subsubsection{Analysis on Consistency Types}\label{section.const_type}
We confirmed that all models exhibited consistent behaviours in \textsc{SIGNAL} case (i.e. higher than 90\%) and close to, or even better than human's performance. On the contrary, the consistency for \textsc{REVERSE} case is considerably lower than that of the \textsc{SIGNAL} case in the majority of cases. The performance for the \textsc{REVERSE} case shows a huge difference with that of humans. The results reveal that the PLMs are sensitive to changes in the order of input sentences.

\subsubsection{Analysis on Model Types}
In our results, the encoder-based models (e.g. $RoBERTa$, $Electra$, $BART$, $T5$) showed better consistency results in all tasks than the generative decoder-based model (e.g. $GPT2$), based on the size of the $base$ model. This implies that the bidirectional encoder \& MLM objective are more effective than the unidirectional decoder \& general LM objective in terms of consistency, which is limited to the classification task. 
Meanwhile, the T5 large model showed the best consistency results for the \textsc{REVERSE} case in all tasks except for a poorly trained case (i.e. $Electra_{small}$ - MRPC task). We conjecture a leading cause is that the $T5$ model used in our experiment was fine-tuned with a multi-task including STS tasks, and this is the result of functioning as paraphrase identification tasks.

\begin{table*}[t!]
	\begin{center}
		\renewcommand{\arraystretch}{1.5}
		\footnotesize{
			\centering{\setlength\tabcolsep{1pt}
		\begin{tabular}{c|c|c|c|c}
		\toprule
		\hline
        Dataset & Type & Input sentence1 & Input sentence2 & Prediction \\
        \hline
        \multirow{2}{*}{RTE} 
        & Original   & \makecell[l]{Sentence1: Microsoft was established \\ in Italy in 1985.} & \makecell[l]{Sentence2: Microsoft was established in 1985.} & entailment \\
        & Signal & \makecell[l]{[Sentence1] Microsoft was established \\ in Italy in 1985.}   &  \makecell[l]{[Sentence2] Microsoft was established in 1985.} & not\_entailment \\
        \hline
        
        \multirow{2}{*}{MRPC} 
        & Original & \makecell[l]{Sentence1: Spinnaker employs roughly \\ 83 people ; NetApp employs 2,400.} & \makecell[l]{Sentence2: Spinnaker employs 83 people, \\ most of whom are engineers.} & equivalent \\
        & Reverse  & \makecell[l]{Sentence2: Spinnaker employs 83 people, \\ most of whom are engineers.}  & \makecell[l]{Sentence1: Spinnaker employs roughly \\ 83 people ; NetApp employs 2,400.} & not\_equivalent \\
        \hline
        \multirow{2}{*}{QNLI} 
        & Original & \makecell[l]{Question: With what word was Tesla's \\ sociability described?} & \makecell[l]{Sentence: Tesla was asocial and prone to \\ seclude himself with his work.} & entailment \\
        & Reverse  & \makecell[l]{Sentence: Tesla was asocial and prone to \\ seclude himself with his work.}  & \makecell[l]{Question: With what word was Tesla's \\ sociability described?} & not\_entailment \\
        \hline
        \bottomrule
        \end{tabular}}}
	\caption{Examples of inconsistent predictions of $RoBERTa_{large}$ model.} \label{table.Example_roberta}%
	\end{center}
\end{table*}

\begin{table*}
	\begin{center}
		\renewcommand{\arraystretch}{1.5}
		\footnotesize{
			\centering{\setlength\tabcolsep{2pt}
				\begin{tabular}{p{80mm}|p{80mm}}
					\toprule
                    \multicolumn{2}{c}{\makecell[l]{
                    \textsc{Original Inputs}: mrpc sentence1: "I felt that if I disagreed with Rosie too much I would lose my job," she said. sentence2: \\ Cavender did say : "I felt that if I disagreed with Rosie too much I would lose my job.}}     
                    \\
                    \multicolumn{2}{c}{\makecell[l]{
                    \textsc{Signal Inputs}: mrpc \colorbox{orange}{[sentence1]} "I felt that if I disagreed with Rosie too much I would lose my job," she said. \colorbox{orange}{[sentence2]} \\ Cavender did say : "I felt that if I disagreed with Rosie too much I would lose my job.}}     
                    \\ \midrule
                    \makecell[c]{\textsc{Original Prediction}} & \makecell[c]{\textsc{Signal Prediction}} \\
                    \makecell[c]{equivalent} & \makecell[c]{<extra\_id\_0>.[sentence1] [sentence2] [sent }\\
                    \midrule
                    
                    \multicolumn{2}{c}{\makecell[l]{
                    \textsc{Original Inputs}: mrpc sentence1: The bishop told police he thought he had hit a dog or a cat or that someone had thrown  \\ a rock at his vehicle. sentence2: Bishop O 'Brien, aged 67, had told police he thought he had hit a dog or cat.}}     
                    \\
                    \multicolumn{2}{c}{\makecell[l]{
                    \textsc{Signal Inputs}: mrpc \colorbox{orange}{[sentence1]} The bishop told police he thought he had hit a dog or a cat or that someone had thrown  \\ a rock at his vehicle.  \colorbox{orange}{[sentence2]} Bishop O 'Brien, aged 67, had told police he thought he had hit a dog or cat.}}     
                    \\ \midrule
                    \makecell[c]{\textsc{Original Prediction}} & \makecell[c]{\textsc{Signal Prediction}} \\
                    \makecell[c]{not\_equivalent} & \makecell[c]{<extra\_id\_0>] [sentence2] [sentence3] [sentence}\\
                    \midrule
                    
                    \multicolumn{2}{c}{\makecell[l]{
                    \textsc{Original Inputs}: mrpc sentence1: John Jacoby, the fire department's battalion chief, arrived. sentence2: Fire Chief John \\ Jacoby arrived, joining Moriarty at the top of the embankment.}}     
                    \\
                    \multicolumn{2}{c}{\makecell[l]{
                    \textsc{Signal Inputs}: mrpc \colorbox{orange}{[sentence1]} John Jacoby, the fire department's battalion chief, arrived.  \colorbox{orange}{[sentence2]} Fire Chief John \\ Jacoby arrived, joining Moriarty at the top of the embankment.}}     
                    \\ \midrule
                    \makecell[c]{\textsc{Original Prediction}} & \makecell[c]{\textsc{Signal Prediction}} \\
                    \makecell[c]{not\_equivalent} & \makecell[c]{<extra\_id\_0>, arrived.<extra\_id\_1>'s battalion chief}\\
					\bottomrule	
		\end{tabular}}}
	\caption{Examples of inconsistent predictions of $T5_{large}$ model on the \textsc{Signal} case of MRPC dataset. The changes made in \textsc{Signal} case inputs are in orange.} \label{table.example_t5_signal}%
	\end{center}
\end{table*}

\subsubsection{Analysis on the Model-size}
In the performance comparison of the $large$ model and the $base$ ($small$) model, The large models significantly improved the accuracy in all tasks. They also showed a tendency to increase overall consistency for tasks (MNI, QNLI, QQP) with large data, but this varied depending on models since there were a few cases of decreased consistency in the REVERSE case.
In the other two tasks (RTE, MRPC), rather, the smaller-sized model showed better consistency performance. We conjecture that over-fitting is likely to occur when fine-tuning a large model with small data, causing a decrease in generalisation ability which makes a model generate incorrect predictions for the perturbed inputs.

These results suggest that a model with high accuracy cannot always fully understand the language, and raise the need to evaluate the model's performance from other lenses such as consistency.


\subsubsection{Analysis on the T5 Models}\label{section:t5_analysis}
Compared to other models that showed robust and high performance in the \textsc{SIGNAL} case of all tasks, 
the performance of $T5$ models in the \textsc{SIGNAL} case falls short of expectation; evaluation was performed 
with text generated by the $T5$ models. One of the strong reasons is that during the fine-tuning of the $T5$ model, the template of all training data contains the colon, which is the same form with that of our original case, but does not include any brackets. Therefore, our \textsc{SIGNAL} case inputs with brackets became a completely new distribution to the model, and as a result, the desired results (label) were not properly generated.
 Several generated examples of the $T5$ models are provided in Table \ref{table.example_t5_signal}. More examples are available in Table \ref{table.example_t5base_signal} in Appendix.

\begin{table*}[t!]
	\begin{center}
		\renewcommand{\arraystretch}{1.3}
		\footnotesize{
			\centering{\setlength\tabcolsep{4pt}
		\begin{tabular}{c|ccc|ccc|ccc}
		\toprule
		\hline
		\multirow{2}{*}{Model} & 
		\multicolumn{3}{c|}{KorNLI} & \multicolumn{3}{c|}{KLUE-NLI} & \multicolumn{3}{c}{KLUE-STS} \\ 
		& $Acc_{val}$ & $C_{R}$ & $C_{S}$ 
		& $Acc_{val}$ & $C_{R}$ & $C_{S}$ 
		& $Acc_{val}$ & $C_{R}$ & $C_{S}$ 
		\\ \hline
		$KoBERT$ 
		& 85.6 & 55.2 & 96.4 
		& 75.7 & 50.1 & 94.4 
		& 79.2 & 92.3 & 94.2 
		\\
		$KoElectra$ 
		& 86.3 & 53.5 & \textbf{98.0} 
		& 78.6 & 56.3 & 96.3 
		& 72.8 & 93.5 & 97.0 
		\\ \hline
		$KoGPT2$ 
		& 84.0 & 49.1 & 89.1 
		& 64.5 & 50.9 & 88.1 
		& 77.3 & 84.9 & 79.1 
		\\
		$KoBART$ 
		& 85.2 & 54.6 & 97.4 
		& 71.5 & 53.9 & 93.1 
		& 76.9 & 86.1 & 97.3 
		\\
		\hline
		Human
		& \textbf{87.3} & \textbf{94.0} & 96.0 
		& \textbf{86.0} & \textbf{98.0} & \textbf{98.0} 
		& \textbf{88.0} & \textbf{100.0} & \textbf{100.0} 
		\\

		\hline
		\bottomrule
		\end{tabular}}}
	\caption{Results for the consistency evaluation on Korean datasets. $Acc_{val}$ denotes an accuracy on the validation dataset. $C_{R}$ and $C_{S}$ stand for the consistency for \textsc{REVERSE} and \textsc{SIGNAL} cases, respectively.  We trained each model 5 times and recorded the average of each metric. The best values are in bold.} \label{table.kor_experiment}%
	\end{center}
\end{table*}

\subsection{Experiments on Korean Datasets}
The experimental results for Korean datasets are provided in Table \ref{table.kor_experiment}. It is interesting that the results for Korean datasets exhibited a similar trend with those for English datasets. The consistency of the \textsc{SIGNAL} case is considerably higher than that of the \textsc{REVERSE} case. Also, models trained on STS tasks recorded high consistency in both \textsc{REVERSE} and \textsc{SIGNAL} case, while those trained on NLI tasks completely failed in \textsc{REVERSE} case. Finally, encoder-based models, such as KoBERT and KoElectra, generally delivered higher consistency than KoGPT2, which is a decoder-based model. The results indicate that the inconsistency problem of PLMs is not to blame languages but the models themselves.

\begin{table*}[t!]
	\begin{center}
		\renewcommand{\arraystretch}{1.3}
		\footnotesize{
			\centering{\setlength\tabcolsep{4pt}
		\begin{tabular}{c|ccc|ccc|ccc}
		\toprule
		\hline
		\multirow{2}{*}{Model} & 
		\multicolumn{3}{c|}{MNLI} & \multicolumn{3}{c|}{QNLI} & \multicolumn{3}{c}{RTE} \\ 
		& $Acc_{val}$ & $C_{R}$ & $C_{S}$ 
		& $Acc_{val}$ & $C_{R}$ & $C_{S}$ 
		& $Acc_{val}$ & $C_{R}$ & $C_{S}$ 
		\\ \hline
		$RoBERTa_{base}{\text -}Single$ 
        & \textbf{87.2} & 60.3 & \textbf{98.6} 
		& \textbf{92.4} & 75.5 & \textbf{98.5} 
		& 66.7 & 66.4 & 92.2 
		\\
		$RoBERTa_{base}{\text -}Para$ 
		& 85.7 & 65.4 & 98.0 
		& 90.1 & \textbf{76.1} & 97.9 
		& 67.9 & 86.5 & 96.1 
		\\
		$RoBERTa_{base}{\text -}All$ 
		& 85.6 & \textbf{66.7} & 97.9 
		& 90.4 & 72.9 & 98.2 
		& \textbf{72.9} & \textbf{89.0} & \textbf{96.9} 
		
		\\ \hline
		$Electra_{large}{\text -}Single$ 
		& \textbf{90.7} & 63.3 & \textbf{98.6} 
		& \textbf{94.7} & 55.4 &	\textbf{99.0} 
		& 82.2 & 59.2 & 94.8 
		\\
		$Electra_{large}{\text -}Para$ 
		& 90.6 & 65.2 & 98.1 
		& 94.3 & \textbf{58.7} & 98.6 
		& 81.6 & 71.0 & 93.1 
		\\
		$Electra_{large}{\text -}All$ 
		& 90.6 & \textbf{66.0} & 98.1 
		& 94.6 & 55.8 & 98.5 
		& \textbf{86.3} & \textbf{77.8} & \textbf{97.7} 
		
		\\ \hline
		$GPT2_{base}{\text -}Single$ 
		& \textbf{80.4} & 44.0 & 92.2 
		& \textbf{87.4} & 48.9 &	94.1 
		& 61.0 & 62.5 & 83.4 
		\\
		$GPT2_{base}{\text -}Para$ 
		& 77.3 & \textbf{54.1} & \textbf{92.8} 
		& 85.6 & \textbf{58.4} & 93.6 
		& \textbf{63.5} & 65.9 & \textbf{90.1} 
		\\
		$GPT2_{base}{\text -}All$ 
		& 77.6 & 49.5 & 92.0 
		& 86.3 & 51.3 & \textbf{95.9} 
		& 60.2 & \textbf{70.3} & 88.1 
		\\
		\hline
		\bottomrule
		\end{tabular}}}
	\caption{Results for the consistency evaluation on multi-task training. $Acc_{val}$ denotes an accuracy on the validation dataset. $C_{R}$ and $C_{S}$ stands for the consistency for \textsc{REVERSE} and \textsc{SIGNAL} cases, respectively.  We trained each model five times and recorded the average of each metric. The best value is in bold.} \label{table.multitask}%
	\end{center}
\end{table*}

\section{Multi-task Training with STS task Improves Consistency}

From the earlier experiments, we observed that models fine-tuned on STS tasks exhibit high consistency. Also, $T5$ models, which are trained in multi-task settings including STS tasks, are more consistent in \textsc{REVERSE} case, compared to the single-task trained models. Based on these results, we made the following assumption:

\medskip 
\noindent \textbf{Assumption.} Using STS tasks as an auxiliary training objective contributes to the improvement of consistency.

\medskip 
We conduct an additional experiment to verify the aformentioned assumption.

\subsection{Experimental Design}
To train a model on multiple datasets simultaneously, we used the structure of MT-DNN \cite{mtdnn}, which shares the encoder but have individual classifiers for each task. We train the following two variants.

\paragraph{Multitask-Para} A model is trained on three tasks: one for the main task (e.g. MNLI, QNLI, or RTE), and the others for auxillary STS tasks (i.e. QQP and MRPC). 

\paragraph{Multitask-All} A model is trained on all five tasks. We implement this variant as a control group to check whether including none-STS datasets as auxiliary tasks could also improve consistency.

As backbone model candidates, we select $RoBERTa_{base}$ and $Electra_{large}$ because $Electra_{small}$ is not trained well on MRPC dataset. We apply the CALM framework to these models: first train the models under the original input format and then measure consistency for the \textsc{REVERSE} and \textsc{SIGNAL} cases. For training, we use the same training options described in section \ref{section.training_detail}.

\subsection{Results}
The experimental results are summarised in Table \ref{table.multitask}. For all types of models, accuracy and consistency of \textsc{SIGNAL} case in MNLI \& QNLI tasks decreased slightly compared to the original model ($Single$), but the differences were marginal. Meanwhile, for the \textsc{REVERSE} case, multitask models ($Para$ \& $ALL$) showed significant improvement of consistency in all tasks by 13\% on average. 
In MNLI task, the multitask $ALL$ model of encoder-based models ($RoBERTa$, $Electra$) showed higher consistency in \textsc{REVERSE} case than that of the $Para$ model, but there was no significant difference. Rather, in the decoder-based model ($GPT2$), the $Para$ model showed the highest consistency in the \textsc{REVERSE} case, showing a significant difference with the $ALL$ model. In the QNLI task, the $PARA$ models showed a higher consistency improvement in the \textsc{REVERSE} case than the $ALL$ models, regardless of the model type. These results indicate that consistency can be improved regardless of the task type by only using the STS tasks as the auxiliary training objective.

On the contrary, in the RTE task, as consistent with the results in MT-DNN \cite{mtdnn}, the accuracy of encoder-based models improved when the size of the training data was small. Additionally, we confirmed that the consistency for the \textsc{REVERSE} and \textsc{SIGNAL} cases also improved dramatically. Especially, although both the accuracy and consistency performance of the $ALL$ models were higher that those of the $Para$ models, even leveraging only paraphrase data showed a significant improvement in the \textsc{REVERSE} case. In the decoder-based model, the accuracy and consistency for the \textsc{SIGNAL} case of the $Para$ model were higher than the $All$ model, indicating that encoder-based and decoder-based models certainly work differently. 

Our work is distinguished from the previous studies \cite{ribeiro-etal-2019-red, asaihajishirzi2020logic, elazar2021measuring} for consistency improvement in that there is no need for augmentation of customised data for a specific task. We figured out that multitask learning with a well-known STS tasks can improve consistency.

\section{Summary and Outlook} 
Unsupervised pretraining and then fine-tuning framework has become a predominant approach in NLP. The outstanding performance of PLMs on diverse NLP tasks is indisputable. However, their language understanding ability is questionable considering the evidence presented in many recent works.

Consistency is a highly desirable property that a good language understanding model should possess to achieve human-level language understanding capability. In this paper, we propose a simple framework named \textbf{CALUM} that measures consistency by adding perturbations that preserve identical meaning. Through experiments, we observed two important findings. First, PLMs exhibit inconsistent behaviours, i.e., they are prone to make different decisions for inputs conveying the same meaning. Second, we revealed that leveraging STS tasks as an auxiliary training objective is of benefit to improve consistency. Our findings suggest that focusing on semantic meaning could be a key for training high-level language understanding models.

Although we confirmed the effect of leveraging STS tasks, it is still far from reaching human-level consistency. In our experiments, we only used paraphrase data for improving consistency. As future work, leveraging other meaning-understanding datasets, such as negated sentences \cite{kassner2020negated} and word definition \cite{senel2021does}, for better consistency would be an interesting research direction.

\clearpage
\bibliography{emnlp2021}
\bibliographystyle{acl_natbib}
\clearpage

\appendix

\section{Appendix}
\label{sec:appendix}
\subsection{Hyper Parameter Search}
Based on the work of \citet{BERT}, we investigated the following range of possible values to decide the hyperparameters for fine-tuning:

\begin{itemize}
    \item Batch size: 32, 64, 128
    \item Learning rate: 5\textit{e}-4, 1\textit{e}-5, 5\textit{e}-6,  
\end{itemize}

Datasets with a large amount of training data, e.g. MNLI and QQP, were not sensitive to the hyperparameter values. Therefore, we selected hyperparameter values that generally performs well on small-sized datasets for all models.

\begin{table*}[t!]
	\begin{center}
		\renewcommand{\arraystretch}{1.5}
		\footnotesize{
			\centering{\setlength\tabcolsep{1pt}
		\begin{tabular}{c|c|c|c|c}
		\toprule
		\hline
        Dataset & Type & Input sentence1 & Input sentence2 & Prediction \\
        \hline
        \multirow{2}{*}{RTE} 
        & Original   & \makecell[l]{Sentence1: These folk art traditions have\\ been preserved for hundreds of years.} & \makecell[l]{Sentence2: Indigenous folk art is preserved.} & entailment \\
        & Signal & \makecell[l]{[Sentence1] These folk art traditions have \\been preserved for hundreds of years.}   &  \makecell[l]{[Sentence2] Indigenous folk art is preserved.} & not\_entailment \\
        \hline

        \multirow{2}{*}{MRPC} 
        & Original & \makecell[l]{Sentence1: The initial report was made to \\ Modesto Police December 28.} & \makecell[l]{Sentence2: It stems from a Modesto police \\ report.} & equivalent \\
        & Reverse & \makecell[l]{Sentence2: It stems from a Modesto police \\ report.} & \makecell[l]{Sentence1: The initial report was made to \\ Modesto Police December 28.} & \makecell[l]{not\_equivalent} \\
        \hline
        
        \multirow{2}{*}{QNLI} 
        & Original & \makecell[l]{Question: What is essential for the successful \\ execution of a project?} & \makecell[l]{Sentence: For the successful execution of a \\ project, effective planning is essential.} & entailment \\
        & Reverse & \makecell[l]{Sentence: For the successful execution of a \\ project, effective planning is essential.} & \makecell[l]{Question: What is essential for the successful \\ execution of a project?} & not\_entailment \\
        \hline
        \bottomrule
        \end{tabular}}}
        
	\caption{Examples of inconsistent predictions of $Electra_{large}$ model.} \label{table.Example_electra_large}%
	\end{center}
\end{table*}

\begin{table*}[t!]
	\begin{center}
		\renewcommand{\arraystretch}{1.5}
		\footnotesize{
			\centering{\setlength\tabcolsep{1pt}
		\begin{tabular}{c|c|c|c|c}
		\toprule
		\hline
        Dataset & Type & Input sentence1 & Input sentence2 & Prediction \\
        \hline
        \multirow{2}{*}{RTE} 
        & Original   & \makecell[l]{Sentence1: In 1900 Berlin's arterial roads ran \\ across Potsdam Square - Potsdamer Platz.} & \makecell[l]{Sentence2: Postdam Square is located in \\ Berlin.} & not\_entailment \\
        & Reverse & \makecell[l]{Sentence2: Postdam Square is located in \\ Berlin.}   &  \makecell[l]{Sentence1: In 1900 Berlin's arterial roads ran \\ across Potsdam Square - Potsdamer Platz.} & entailment \\
        \hline
        
        \multirow{2}{*}{MRPC} 
        & Original & \makecell[l]{Sentence1: The study is being published \\ today in the journal Science.} & \makecell[l]{Sentence2: Their findings were published \\ today in Science.} & equivalent \\
        & Signal & \makecell[l]{[Sentence1] The study is being published \\ today in the journal Science.} & \makecell[l]{[Sentence2] Their findings were published \\ today in Science.} & \makecell[l]{<extra\_id\_0>...} \\
        \hline
        
        \multirow{2}{*}{QNLI} 
        & Original & \makecell[l]{Question: What fueled Luther's  concept of \\ Christ and His Salvation?} & \makecell[l]{Sentence: His railing against the sale of \\ indulgences was based on it.} & not\_entailment \\
        & Signal & \makecell[l]{[Question] What fueled Luther's  concept of \\ Christ and His Salvation?} & \makecell[l]{[Sentence] His railing against the sale of \\ indulgences was based on it.} & entailment \\
        \hline
        \bottomrule
        \end{tabular}}}
	\caption{Examples of inconsistent predictions of $T5_{large}$ model. The generated output of \textsc{SIGNAL} case in MRPC dataset is `<extra\_id\_0>1]<extra\_id\_1>. [sentence3] The study is being".} \label{table.Example_t5_large}%
	\end{center}
\end{table*}

\begin{table*}
	\begin{center}
		\renewcommand{\arraystretch}{1.5}
		\footnotesize{
			\centering{\setlength\tabcolsep{2pt}
				\begin{tabular}{p{80mm}|p{80mm}}
					\toprule
                    \multicolumn{2}{c}{\makecell[l]{
                    \textsc{Original Inputs}: rte sentence1: At least 50 animals died in a late December avalanche. sentence2: Humans died \\ in an avalanche.}}     
                    \\
                    \multicolumn{2}{c}{\makecell[l]{
                    \textsc{Original Inputs}: rte \colorbox{orange}{[sentence1]} At least 50 animals died in a late December avalanche. \colorbox{orange}{[sentence2]} Humans died \\in an avalanche.}}    
                                      
                    \\ \midrule
                    \makecell[c]{\textsc{Original Prediction}} & \makecell[c]{\textsc{Signal Prediction}} \\
                    \makecell[c]{not\_entailment} & \makecell[c]{<extra\_id\_0> <extra\_id\_1> [sentence1] At least 50\\ animals died in an}\\
                    \midrule
                    \multicolumn{2}{c}{\makecell[l]{
                    \textsc{Original Inputs}: rte sentence1: Microsoft denies that it holds a monopoly. sentence2: Microsoft holds a monopoly power.}}     
                    \\
                    \multicolumn{2}{c}{\makecell[l]{
                    \textsc{Signal Inputs}: rte \colorbox{orange}{[sentence1]} icrosoft denies that it holds a monopoly.  \colorbox{orange}{[sentence2]} Microsoft holds a monopoly power.}}     
                    \\ \midrule
                    \makecell[c]{\textsc{Original Prediction}} & \makecell[c]{\textsc{Signal Prediction}} \\
                    \makecell[c]{not\_entailment} & \makecell[c]{[sentence1] Microsoft denies that it holds a monopol}\\
                    \midrule
                    
                    \multicolumn{2}{c}{\makecell[l]{
                    \textsc{Original Inputs}: rte sentence1: An earthquake has hit the east coast of Hokkaido, Japan, with a magnitude of 7.0 Mw. \\ sentence2: An earthquake occurred on the east coast of Hokkaido, Japan.}}     
                    \\
                    \multicolumn{2}{c}{\makecell[l]{
                    \textsc{Signal Inputs}: rte \colorbox{orange}{[sentence1]} An earthquake has hit the east coast of Hokkaido, Japan, with a magnitude of 7.0 Mw. \\ \colorbox{orange}{[sentence2]} An earthquake occurred on the east coast of Hokkaido, Japan.}}     
                    \\ \midrule
                    \makecell[c]{\textsc{Original Prediction}} & \makecell[c]{\textsc{Signal Prediction}} \\
                    \makecell[c]{entailment} & \makecell[c]{<extra\_id\_0>e<extra\_id\_1>e [sentence2] An earthquake has\\ hit the east}\\
					\bottomrule	
		\end{tabular}}}
	\caption{More examples of inconsistent predictions of $T5_{base}$ model on the \textsc{Signal} case of RTE dataset. The changes made in \textsc{Signal} case inputs are in orange.} \label{table.example_t5base_signal}%
	\end{center}
\end{table*}


\end{document}